\documentclass{article}
\usepackage{times}
\usepackage{epsfig}
\usepackage{graphicx}
\usepackage{amsmath}
\usepackage{amssymb}
\usepackage{times}
\usepackage{epsfig}
\usepackage{graphicx}
\usepackage{amsmath}
\usepackage{amssymb}
\usepackage{xcolor}
\usepackage[ruled,vlined,onelanguage]{algorithm2e}
\usepackage{setspace}
\usepackage{collcell}
\usepackage{xr}
\usepackage{natbib}
\usepackage{multirow}
\usepackage{subcaption}
\usepackage{mathtools}
\usepackage{colortbl,dcolumn}
\usepackage{algorithm2e}
\usepackage{standalone}

\usepackage[pagebackref=true,breaklinks=true,letterpaper=true,colorlinks,bookmarks=false]{hyperref}

\renewcommand{\eqref}[1]{\hyperref[#1]{Eq.\ \ref*{#1}}}
\newcommand{\figref}[1]{\hyperref[#1]{Fig.\ \ref*{#1}}}
\newcommand{\tabref}[1]{\hyperref[#1]{Table\ \ref*{#1}}}
\newcommand{\secref}[1]{\hyperref[#1]{Section\ \ref*{#1}}}
\newcommand{\algoref}[1]{\hyperref[#1]{Algorithm\ \ref*{#1}}}
\newcolumntype{C}[1]{>{\centering}m{#1}}

\newcommand{\std}[1]{ \normalfont \color{darkgray}\footnotesize{$\pm$#1} }

\newcommand{\E}{{\mathbb E}}
\newcommand{\Q}{{\mathbb Q}}

\hypersetup{
    colorlinks=true,
    linkcolor=blue,
    filecolor=magenta,      
    urlcolor=blue,
}

\usepackage{microtype}
\usepackage{booktabs} 
\usepackage{hyperref}

\usepackage[accepted]{icml2020}

\icmltitlerunning{Bayesian active learning for production}
\begin{document}

\twocolumn[
\icmltitle{Bayesian active learning for production, a systematic study and a reusable library}



\icmlsetsymbol{equal}{*}

\begin{icmlauthorlist}
\icmlauthor{Parmida Atighehchian}{equal,eai}
\icmlauthor{Fr\'ed\'eric Branchaud-Charron}{equal,eai}
\icmlauthor{Alexandre Lacoste}{eai}
\end{icmlauthorlist}

\icmlaffiliation{eai}{Element AI, Montr\'eal, Canada}

\icmlcorrespondingauthor{Fr\'ed\'eric Branchaud-Charron}{frederic.branchaud-charron@elementai.com}

\icmlkeywords{Machine Learning, Active Learning, Uncertainty Estimation}

\vskip 0.3in
]



\printAffiliationsAndNotice{\icmlEqualContribution} 

\begin{figure}
    \centering
    \includegraphics[width=0.4\textwidth]{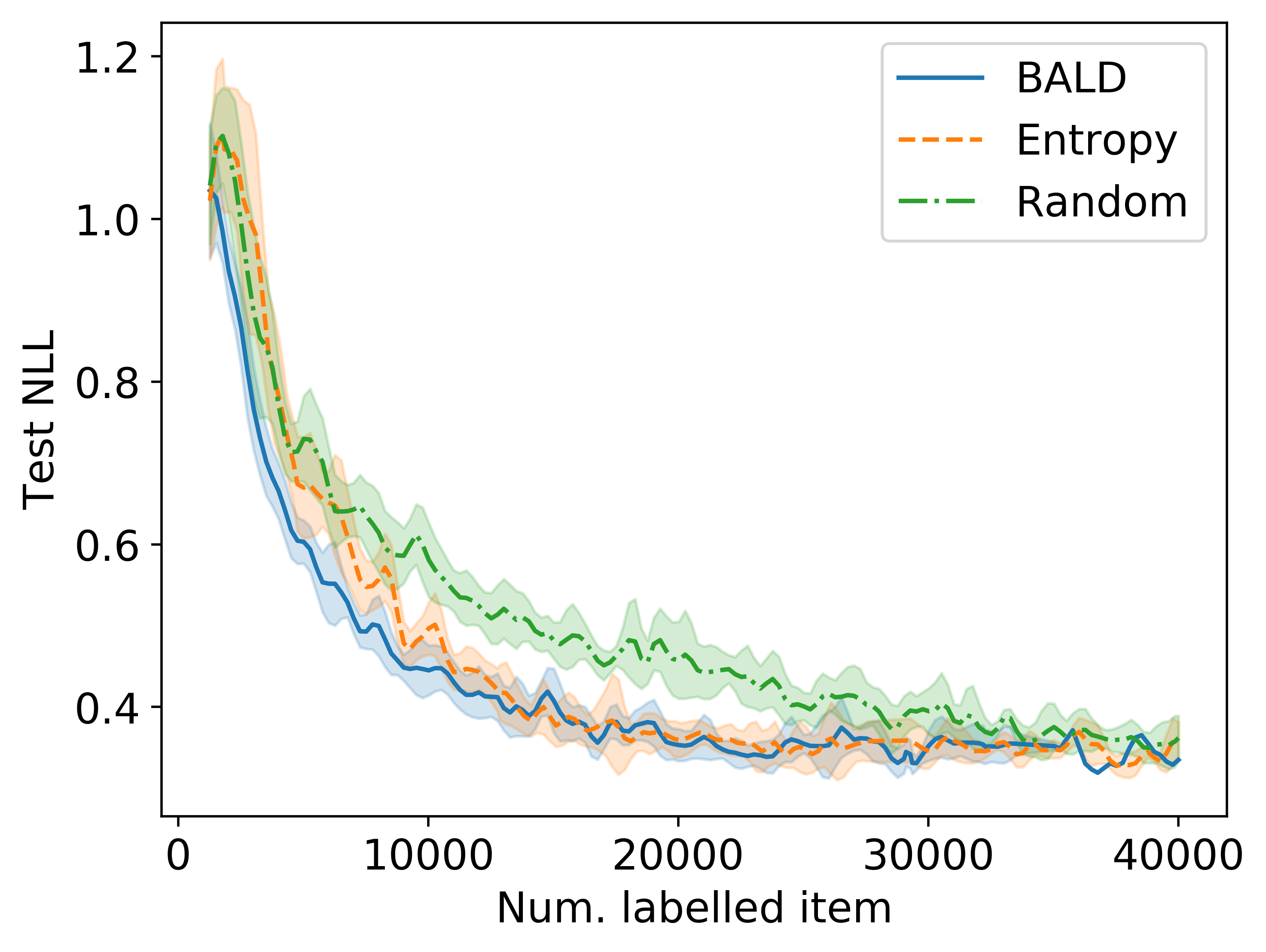}
    \caption{Baselines results on CIFAR10 using MC-Dropout and VGG-16. On an academic dataset, both active learning techniques are competitive.}
    \label{fig:standard}
\end{figure}

\begin{abstract}
\vspace{-0.60cm}
   Active learning is able to reduce the amount of labelling effort by using a machine learning model to query the user for specific inputs.
   While there are many papers on new active learning techniques, these techniques rarely satisfy the constraints of a real-world project. In this paper, we analyse the main drawbacks of current active learning techniques and we present approaches to alleviate them. We do a systematic study on the effects of the most common issues of real-world datasets on the deep active learning process: model convergence, annotation error, and dataset imbalance. We derive two techniques that can speed up the active learning loop such as partial uncertainty sampling and larger query size. Finally, we present our open-source Bayesian active learning library, BaaL.
\end{abstract}


\section{Introduction}
The amount of data readily available for machine learning has exploded in recent years. 
However, for data to be used for deep learning models, labelling is often a required step. 
A common problem when labelling new datasets is the required human effort to perform the annotation. In particular, tasks that require particular domain knowledge such as medical imaging, are expensive to annotate. To solve this, active learning (AL) has been proposed to only label the core set of observations useful for training.

While the active learning field includes many approaches \citep{kirsch2019batchbald, tsymbalov2019deeper, beluch2018power, maddox2019simple}, these methods are often not scalable to large datasets or too slow to be used in a more realistic environment e.g. in a production setup. In particular, active learning applied to images or text requires the usage of deep learning models which are slow to train and themselves require a noticeable huge amount of data to be effective \citep{imagenet_cvpr09, abu2016youtube}.

Furthermore, deep learning models require carefully tuned hyperparameters to be effective. In a research environment, one can fine-tune and perform hyperparameter search to find the optimal combination that gives the biggest reduction in labelling  effort. In a real-world setting, the hyperparameters are set at the beginning with no guarantee for the outcome.

Finally, in a real-world setup, the data is often not cleaned nor balanced. In particular, studies have shown that humans are far from perfect when labelling and the problem is even worse when using crowd-sourcing \citep{ipeirotis2010quality, allahbakhsh2013quality}.

Our contributions are three-fold. We perform a systematic study on the effect of the most common pathologies found in real-world datasets on active learning. Second, we propose several techniques that make active learning suitable for production. Finally, we present a case study of using active learning on a real-world dataset. 

In addition, we present our freely available Bayesian active learning library, BaaL\footnote{\url{https://github.com/ElementAI/baal}} which provides all the necessary set up and tools for active learning experiments at any scale.

\section{Problem setting}

We consider the problem of supervised learning where we observe a dataset made of $n$ pairs $D_L= \{(x_i, y_i)\}_i^{N}$ and our goal is to estimate a prediction function $p(y\mid x)$. In addition, we have $N'$ observations without label $D_U = \{x_i\}_i^{N'}$. More specifically, we consider the problem of active learning where the algorithm is summarized in \algoref{algo:al}.

\begin{algorithm}
 \KwData{$D = \{x_0, \ldots, x_n\}$}
 \KwResult{$D_L = \{(x_0, y_0), \ldots\}$}
 ~~$D_L \leftarrow$ Label randomly B points\;\\
 $D_U \leftarrow D \setminus D_L$ \;\\
\While{labelling budget is available}{
  Train model to convergence on $D_L$\;\\
  Compute uncertainty $U(x), \text{for all } x \in D_U$\;\\
  Label top-\textit{k} most uncertain samples\;
 }
 \caption{\textbf{Active learning process.} For batch active learning, the algorithm train a model on $D_L$ before estimating the uncertainty on the pool $D_U$, the most uncertain samples are labelled by a human before restarting the loop.}
 \label{algo:al}
\end{algorithm}

\section{Background}\label{background}
Active learning has received a lot of attention in the past years, specifically on classification tasks \citep{gal2017deep}. However, some work has been done on segmentation \citep{kendall2017uncertainties}, localization \citep{miller2019evaluating}, natural language processing \citep{siddhant2018deep}, and time series \citep{peng2017acts}. In this paper, we focus our attention on image classification. 

\paragraph{Bayesian active Learning}
Current state-of-the-art techniques used in active learning relies on uncertainty estimation to perform queries~\citep{gal2017deep}.
A common issue highlighted in \citet{tsymbalov2019deeper, kirsch2019batchbald} is the need to retrain the model and recompute the uncertainties as often as possible. Otherwise, the next samples to be selected may be too similar to previously annotated samples. This is problematic due to the long training time of deep-learning models as well as the expensive task of uncertainty estimation.
\citet{tsymbalov2019deeper, houlsby2011bayesian} and \citet{Wilson2015DeepKL} proposed solutions to this issue, are memory expensive and time-consuming when used on large input size or large datasets.
In reality, due to the large cost of inference and retraining for large scale datasets, it is not feasible to recompute the uncertainties in a timely fashion. In consequence, multiple samples are annotated between retraining. We call this framework batch active learning.

Machine leanings algorithms can suffer from two types of uncertainties \citep{kendall2017uncertainties}:

1) \emph{Aleatoric Uncertainty}, the uncertainty intrinsic to the data, which cannot be explained with more samples. This is due to e.g.: errors during labelling, occlusion, poor data acquisition ,or when two classes are highly confused. 

2) \emph{Epistemic Uncertainty}, the uncertainty about the underlying model. Obtaining more samples will provide more information about the underlying model and reduce the amount of epistemic uncertainty. Crucially, some samples are more informative than others. 

\paragraph{Uncertainty estimations} 
Computing the uncertainty of deep neural networks is crucial to many applications from medical imaging to loan application. Unfortunately, deep neural networks are often overconfident as they are not designed to provide calibrated predictions \citep{scalia2019evaluating,gal2016uncertainty}. Hence, researchers proposed new methods to get a trustful estimation of the epistemic uncertainty such as MC-Dropout \citep{gal2016dropout}, Bayesian neural networks \citep{blundell2015weight} or Ensembles. More recently, \citet{wilson2020bayesian} proposed to combine variational inference and ensembles. While this approach is state-of-the-art, it is far too computationally expensive to be used in the industry.


In this paper, we will use MC-Dropout \citep{gal2016dropout}. In this technique proposes the Dropout layers are kept activated at test time to sample from the posterior distribution. Hence, this method can be used on any architecture that uses Dropout which makes it usable on a wide range of applications.


\paragraph{Acquisition functions}
Many heuristics have been proposed to extract the uncertainty value from the stochastic prediction sampling. We define  Monte-Carlo sampling from the posterior distribution $p(w\mid D)$ as:
\begin{equation*}
    p_t(y \mid x) = p(y \mid x, w_t), t \in \{1 \ldots T\}, w_t \sim p(w\mid D)
\end{equation*}

where $T$ is the number of Monte-Carlo samples. 
We compute the Bayesian model average, $\hat p(y \mid x) = \frac{1}{T} \sum_t^T p_t(y \mid x)$. When highly uncertain, $\hat p(y \mid x)$ will be close to a uniform distribution. A naive approach to estimate the uncertainty is to compute the entropy of this distribution.



A more sophisticated approach is BALD~\citep{houlsby2011bayesian}, which estimates the epistemic uncertainty by computing the mutual information between the model posterior distribution and the prediction:
\begin{equation*}
    I(y, w \mid x, D_L) = H[y \mid x, D_L] - \E_{p(w \mid D_L)}(H[y \mid x, w]).
\end{equation*}

BALD compares the entropy of the mean estimator to the entropies of all estimators. The result is high when there are high disagreements between predictions, which addresses the overconfidence issue in deep learning models.



\section{Experiments}

In this paper, we want to demonstrate the usability of active learning in a real-world scenario. First, we analyze the effect of common pathologies in deep learning on active learning. Secondly, a common issue in active learning is the time required between steps in the active learning loop. As stated by \citet{kirsch2019batchbald}, retraining as soon as possible is crucial to obtain decorrelated samples. We investigate if a) this is the case in large-scale datasets and b) what can we do to make this faster. Implementation details can be found in Annex. Baselines for all  acquisition functions can be found in \figref{fig:standard}.



\subsection{Pathologies}\label{pathologies}
In this section, we verify if common pathologies in deep learning hold for active learning. Problems such as annotation error or  model convergence 
may be hurtful to the procedure and are often overlooked in the literature. In particular, due to the small amount of annotated data, models are more at risk than when they are trained on large datasets.

\paragraph{Effect of annotation error}
While standard datasets are of good quality, humans are far from perfect and will produce errors when labelling. This is especially true when using crowdsourcing \citep{allahbakhsh2013quality}.
Because active learning relies on the training data to train a model and there are only a few labelled samples, we make the hypothesis that active learning would be highly sensitive to noise.

To confirm this hypothesis, we introduce noise by corrupting $\lambda \%$ of the labels. We test our hypothesis on CIFAR10~\citep{krizhevsky2009learning}. In \tabref{fig:labl_noise}, we can assess that depending on $\lambda$, the active learning procedure is highly affected by labelling noise. Furthermore, when we compare to random selection, the gain of using active learning decreases when noise is involved, but it is still useful.

\begin{table}
    \centering
    {\small
    \begin{tabular}{llll}
\toprule
Dataset size &            5000 &           10000 &           20000 \\
\midrule
     $\lambda$=               0 &                 &                 \\
     \hline
        BALD &  0.65 \std{0.01} &  0.53 \std{0.01} &  0.43 \std{0.02} \\
     Entropy &  0.68 \std{0.03} &  0.52 \std{0.02} &  0.43 \std{0.03} \\
      Random &  0.71 \std{0.02} &  0.58 \std{0.02} &  0.47 \std{0.01} \\
      \hline
     $\lambda$=            0.05 &                 &                 \\
     \hline
        BALD &  0.72 \std{0.02} &  0.57 \std{0.01} &  0.43 \std{0.02} \\
     Entropy &  0.72 \std{0.02} &  0.54 \std{0.02} &  0.41 \std{0.01} \\
      Random &  0.73 \std{0.03} &  0.61 \std{0.03} &  0.51 \std{0.02} \\
      \hline
     $\lambda$ =             0.1 &                 &                 \\
     \hline
        BALD &  0.78 \std{0.03} &  0.62 \std{0.01} &  0.48 \std{0.01} \\
     Entropy &  0.71 \std{0.02} &  0.57 \std{0.01} &  0.44 \std{0.02} \\
      Random &  0.76 \std{0.02} &  0.64 \std{0.02} &  0.54 \std{0.01} \\
\bottomrule
\end{tabular}}

    \caption{Effect of annotation error on active learning by randomly shuffling $\lambda$\% labels. The test log-likelihood is averaged over 5 runs.}
    \label{fig:labl_noise}
\end{table}

\paragraph{Effect of model convergence}

Because we have no control over the training regime at each time step, it is hard to train the model to an optimal solution. With fully annotated datasets, we can fine-tune our training setup with hyper-parameter search or train for days at a time. In a production environment, we are limited in our ability to best train the model. 
In consequence, the model may be under or overfitted to the current dataset and provide flawed uncertainty estimations. 

\begin{figure}
    \centering
    \includegraphics[width=0.35\textwidth]{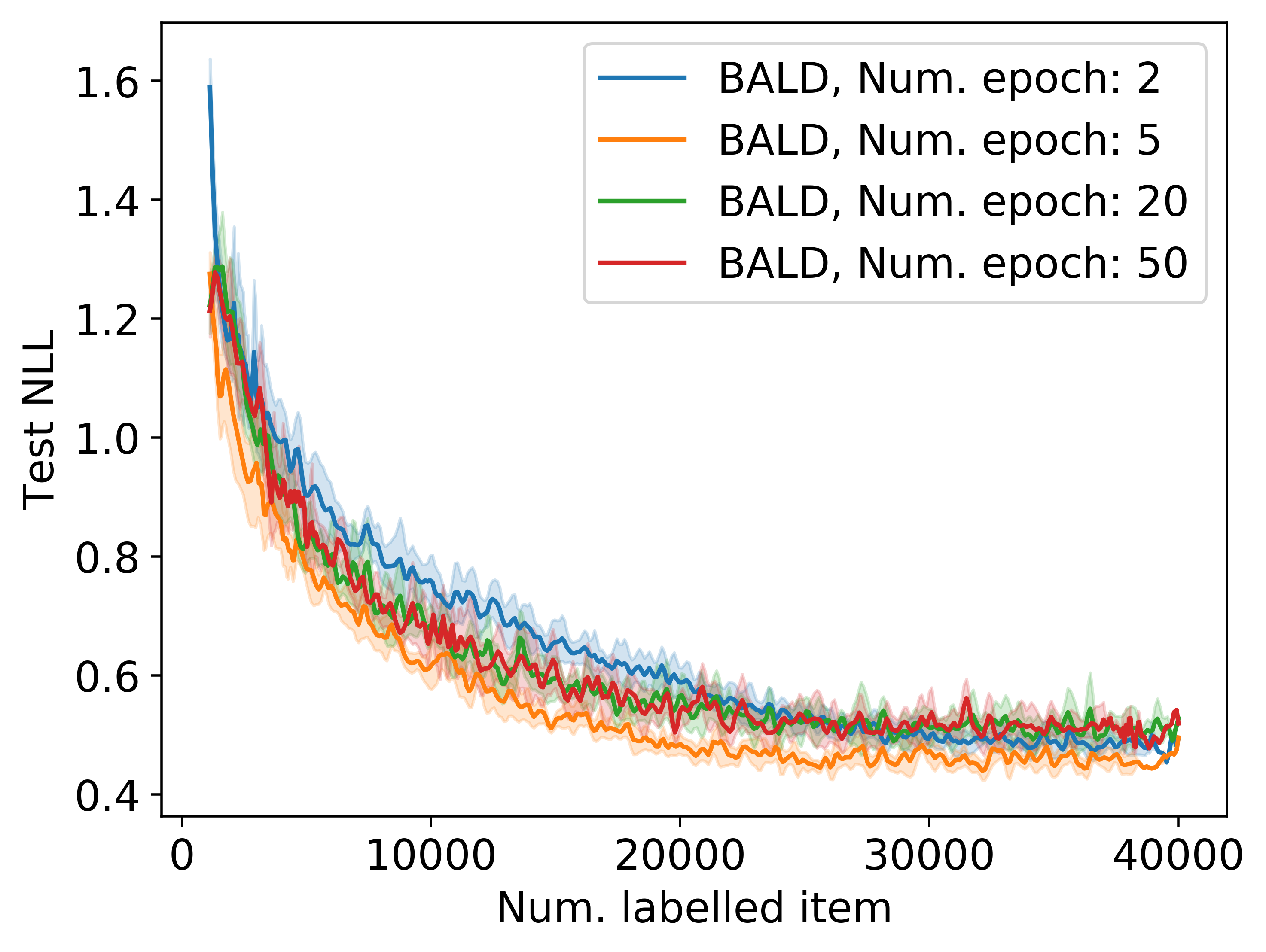}
    \caption{Effect of different training schedules. By comparing overfitted and underfitted models, we assess the impact of uncertainty quality on active learning. Performance averaged over 5 runs.}
    \label{fig:convergence}
\end{figure}

To confirm our hypothesis, we vary the number of epochs the model is trained for. As seen in \figref{fig:convergence}, underfitted models are highly affected while overfitted models suffer, but are still performant. This is due to a poor fit of the model that lead to a wrong estimation of the model uncertainty. 
In Annex, we present the difference in performance between BALD and Random.

In this section, we investigated the effect of two common deep learning pathologies in active learning. In summary, prior knowledge on the quality of the annotations and on how long to train the model, could help using active learning. 

\subsection{Efficient techniques for active learning}\label{techniques}

An important problem with current active learning pipelines is the delay between active learning steps. 
To make active learning efficient, we propose several techniques that maintain performance while speeding up the training or inference phases.

\paragraph{Query size}
An important hyper-parameter in batch active learning is to decide how many samples should be labelled at each active learning step \citep{gal2017deep, tsymbalov2019deeper}. 
In a real-world scenario, we can't ask the annotation team to wait between steps especially in a crowd-sourcing environment. 
Therefore, there needs to be a configuration where we could benefit from a good uncertainty estimation quality and a reasonable runtime. We present our findings in \tabref{fig:query_size} where we tested this approach on CIFAR10. 
From our results, the query size does decreases performances, especially at 10,000 labels where the gap between BALD and Entropy is thinner as the query size grows.

\begin{table}
{\small
\begin{tabular}{llll}
\toprule
    Dataset size &          5000 &         10000 &         20000 \\
\midrule
\textbf{Random} &  0.71 \std{0.03} &  0.54 \std{0.01} &  0.42 \std{0.05} \\
\hline
   $\Q$=50 &                  &                  &                  \\
   \hline
            BALD &  0.59 \std{0.01} &  0.46 \std{0.05} &  0.34 \std{0.02} \\
         Entropy &  0.69 \std{0.06} &  0.55 \std{0.11} &  0.34 \std{0.00} \\
          \hline
  $\Q$=250 &                  &                  &                  \\
  \hline
            BALD &  0.61 \std{0.03} &  0.43 \std{0.01} &  0.35 \std{0.03} \\
         Entropy &  0.67 \std{0.05} &  0.49 \std{0.04} &  0.35 \std{0.00} \\
          \hline
  $\Q$=500 &                  &                  &                  \\
  \hline
            BALD &  0.61 \std{0.07} &  0.42 \std{0.02} &  0.36 \std{0.01} \\
         Entropy &  0.61 \std{0.07} &  0.47 \std{0.00} &  0.37 \std{0.00} \\
          \hline
 $\Q$=2000 &                  &                  &                  \\
 \hline
            BALD &  0.77 \std{0.05} &  0.53 \std{0.03} &  0.37 \std{0.03} \\
         Entropy &  0.87 \std{0.01} &  0.52 \std{0.07} &  0.35 \std{0.01} \\
\bottomrule
\end{tabular}}

    \centering
    \caption{Effect of increasing the query size $\Q$ on CIFAR10. Performance averaged over 5 runs. BALD is weaker when used with a large query size, making Entropy competitive.}
    \label{fig:query_size}
\end{table}

\paragraph{Limit pool size}

The most time-consuming part of active learning is the uncertainty estimation step. In particular, this step is expensive when using techniques that require Monte-Carlo sampling such as MC-Dropout or Bayesian neural networks. 
Of course, this problem is embarrassingly parallel, but for low-budget deployment, one has not access to the resources required to parallelize this task cheaply.
A simple idea to solve this is to randomly select unlabelled samples from the pool instead of using the entire pool. We test this idea by varying the number of samples selected for uncertainty estimation. 
From our experiments (figure in Annex), we show that the performance is not affected when using less than 25\% of the pool. By doing this, we can speed up this phase by a factor of 3.

In this section, we proposed two approaches to make active learning usable in production. First, we can increase the query size higher than previously used in the literature. Second, we can select the next batch using a small subset of the pool.


\section{Case study: Mio-TCD}

Few datasets have been proposed to mimic a "real-world" situation where the dataset suffers from labelling noise, duplicates, or imbalanced datasets. Mio-TCD \citep{luo2018mio} has been recently proposed to showcase these issues. The dataset contains 500,000 samples split into 11 classes with heavy class imbalance. For example, the training set contains 50\% \textit{Cars} and 20\% \textit{Background}. 

\paragraph{Benefits of active learning.}
As shown in \citet{gal2017deep} (and further in Annex), active learning helps when used on imbalanced data. We can verify this, by comparing the performance of underrepresented classes in Mio-TCD. From the current leader board, we can select two difficult classes: \textit{Single-Unit Truck, and Bicycle}. We use the same setup as before, but limit the size of the pool to 20,000 samples. 



\begin{figure}
    \centering
    \includegraphics[width=0.5\textwidth]{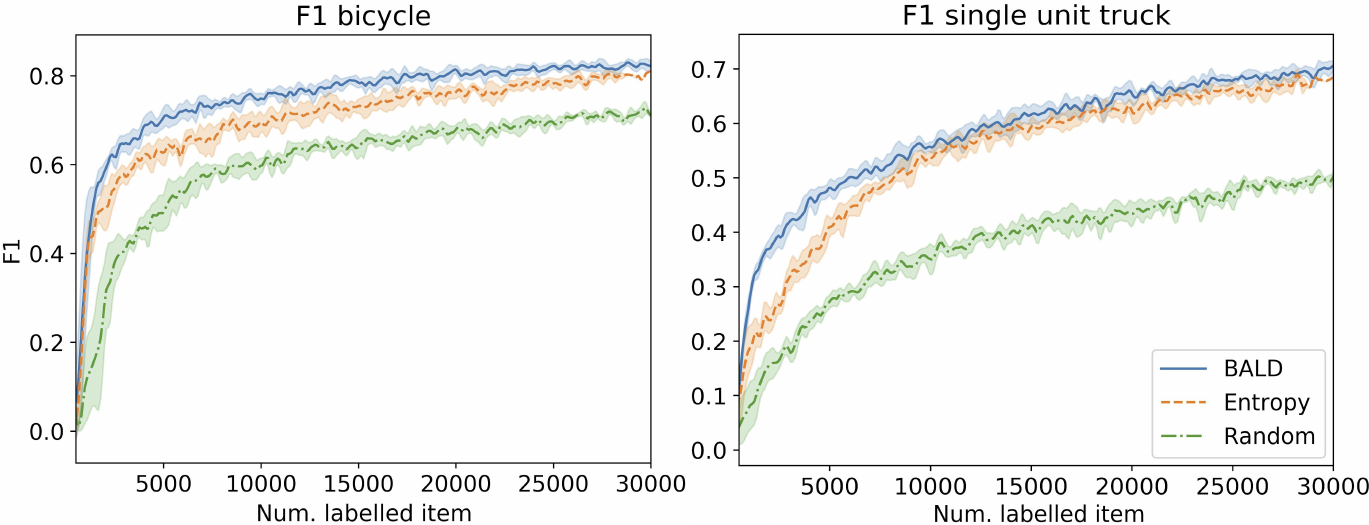}
    \caption{Performance of different active learning procedures on Mio-TCD. While any active learning method is strong against random, BALD is especially strong at the beginning of the labelling process.  Performance averaged over 5 runs.}
    \label{fig:mioclasses}
\end{figure}

In \figref{fig:mioclasses}, we present the F1 scores for the two most difficult classes. One can clearly see the impact of using active learning in this setting. With active learning, underrepresented classes are quickly selected and get decent performance. In addition, the performance for the most populous class \textit{Cars} stays similar across acquisition functions (figure in Annex). 


This experiment shows that using active learning on non-academic datasets is highly beneficial and the need for the active learning community to use new benchmarks to compare methods.

\section{Conclusion}

In this paper, we have investigated the impact of uncleaned data on deep active learning models used for active learning. We also propose several techniques to make active learning usable in a real-world environment. Subsequently, we test our findings on a real-world dataset Mio-TCD, showing that active learning can be used in this setting.  As a result of this study, we introduce our newly released Bayesian active learning library which can be useful to both researchers and developers (see Annex).

In summary, we show that active learning can be used in a production setting on real data with success. We hope this work can fasten the application of active learning on real-world projects and improve the quality of annotation by getting more information per sample. Interesting areas of research include the study of the interaction between the human and the machine during a labelling task.

\clearpage
\bibliography{example_paper}

\begin{thebibliography}{26}
\providecommand{\natexlab}[1]{#1}
\providecommand{\url}[1]{\texttt{#1}}
\expandafter\ifx\csname urlstyle\endcsname\relax
  \providecommand{\doi}[1]{doi: #1}\else
  \providecommand{\doi}{doi: \begingroup \urlstyle{rm}\Url}\fi

\bibitem[Abu-El-Haija et~al.(2016)Abu-El-Haija, Kothari, Lee, Natsev, Toderici,
  Varadarajan, and Vijayanarasimhan]{abu2016youtube}
Abu-El-Haija, S., Kothari, N., Lee, J., Natsev, A.~P., Toderici, G.,
  Varadarajan, B., and Vijayanarasimhan, S.
\newblock Youtube-8m: A large-scale video classification benchmark.
\newblock In \emph{arXiv:1609.08675}, 2016.
\newblock URL \url{https://arxiv.org/pdf/1609.08675v1.pdf}.

\bibitem[Allahbakhsh et~al.(2013)Allahbakhsh, Benatallah, Ignjatovic,
  Motahari-Nezhad, Bertino, and Dustdar]{allahbakhsh2013quality}
Allahbakhsh, M., Benatallah, B., Ignjatovic, A., Motahari-Nezhad, H.~R.,
  Bertino, E., and Dustdar, S.
\newblock Quality control in crowdsourcing systems: Issues and directions.
\newblock \emph{IEEE Internet Computing}, 17\penalty0 (2):\penalty0 76--81,
  2013.

\bibitem[Beluch et~al.(2018)Beluch, Genewein, N{\"u}rnberger, and
  K{\"o}hler]{beluch2018power}
Beluch, W.~H., Genewein, T., N{\"u}rnberger, A., and K{\"o}hler, J.~M.
\newblock The power of ensembles for active learning in image classification.
\newblock In \emph{Proceedings of the IEEE Conference on Computer Vision and
  Pattern Recognition}, pp.\  9368--9377, 2018.

\bibitem[Blundell et~al.(2015)Blundell, Cornebise, Kavukcuoglu, and
  Wierstra]{blundell2015weight}
Blundell, C., Cornebise, J., Kavukcuoglu, K., and Wierstra, D.
\newblock Weight uncertainty in neural networks.
\newblock \emph{arXiv preprint arXiv:1505.05424}, 2015.

\bibitem[Deng et~al.(2009)Deng, Dong, Socher, Li, Li, and
  Fei-Fei]{imagenet_cvpr09}
Deng, J., Dong, W., Socher, R., Li, L.-J., Li, K., and Fei-Fei, L.
\newblock {ImageNet: A Large-Scale Hierarchical Image Database}.
\newblock In \emph{CVPR09}, 2009.

\bibitem[Gal(2016)]{gal2016uncertainty}
Gal, Y.
\newblock Uncertainty in deep learning.
\newblock \emph{University of Cambridge}, 1:\penalty0 3, 2016.

\bibitem[Gal \& Ghahramani(2016)Gal and Ghahramani]{gal2016dropout}
Gal, Y. and Ghahramani, Z.
\newblock Dropout as a bayesian approximation: Representing model uncertainty
  in deep learning.
\newblock In \emph{international conference on machine learning}, pp.\
  1050--1059, 2016.

\bibitem[Gal et~al.(2017)Gal, Islam, and Ghahramani]{gal2017deep}
Gal, Y., Islam, R., and Ghahramani, Z.
\newblock Deep bayesian active learning with image data.
\newblock In \emph{Proceedings of the 34th International Conference on Machine
  Learning-Volume 70}, pp.\  1183--1192. JMLR. org, 2017.

\bibitem[Houlsby et~al.(2011)Houlsby, Husz{\'a}r, Ghahramani, and
  Lengyel]{houlsby2011bayesian}
Houlsby, N., Husz{\'a}r, F., Ghahramani, Z., and Lengyel, M.
\newblock Bayesian active learning for classification and preference learning.
\newblock \emph{arXiv preprint arXiv:1112.5745}, 2011.

\bibitem[Ipeirotis et~al.(2010)Ipeirotis, Provost, and
  Wang]{ipeirotis2010quality}
Ipeirotis, P.~G., Provost, F., and Wang, J.
\newblock Quality management on amazon mechanical turk.
\newblock In \emph{Proceedings of the ACM SIGKDD workshop on human
  computation}, pp.\  64--67, 2010.

\bibitem[Kendall \& Gal(2017)Kendall and Gal]{kendall2017uncertainties}
Kendall, A. and Gal, Y.
\newblock What uncertainties do we need in bayesian deep learning for computer
  vision?
\newblock In \emph{Advances in neural information processing systems}, pp.\
  5574--5584, 2017.

\bibitem[Kirsch et~al.(2019)Kirsch, van Amersfoort, and
  Gal]{kirsch2019batchbald}
Kirsch, A., van Amersfoort, J., and Gal, Y.
\newblock Batchbald: Efficient and diverse batch acquisition for deep bayesian
  active learning, 2019.

\bibitem[Krawczyk(2016)]{krawczyk2016learning}
Krawczyk, B.
\newblock Learning from imbalanced data: open challenges and future directions.
\newblock \emph{Progress in Artificial Intelligence}, 5\penalty0 (4):\penalty0
  221--232, 2016.

\bibitem[Krizhevsky et~al.(2009)Krizhevsky, Hinton,
  et~al.]{krizhevsky2009learning}
Krizhevsky, A., Hinton, G., et~al.
\newblock Learning multiple layers of features from tiny images.
\newblock 2009.

\bibitem[Luo et~al.(2018)Luo, Branchaud-Charron, Lemaire, Konrad, Li, Mishra,
  Achkar, Eichel, and Jodoin]{luo2018mio}
Luo, Z., Branchaud-Charron, F., Lemaire, C., Konrad, J., Li, S., Mishra, A.,
  Achkar, A., Eichel, J., and Jodoin, P.-M.
\newblock Mio-tcd: A new benchmark dataset for vehicle classification and
  localization.
\newblock \emph{IEEE Transactions on Image Processing}, 27\penalty0
  (10):\penalty0 5129--5141, 2018.

\bibitem[Maddox et~al.(2019)Maddox, Izmailov, Garipov, Vetrov, and
  Wilson]{maddox2019simple}
Maddox, W.~J., Izmailov, P., Garipov, T., Vetrov, D.~P., and Wilson, A.~G.
\newblock A simple baseline for bayesian uncertainty in deep learning.
\newblock In \emph{Advances in Neural Information Processing Systems}, pp.\
  13132--13143, 2019.

\bibitem[Miller et~al.(2019)Miller, Dayoub, Milford, and
  S{\"u}nderhauf]{miller2019evaluating}
Miller, D., Dayoub, F., Milford, M., and S{\"u}nderhauf, N.
\newblock Evaluating merging strategies for sampling-based uncertainty
  techniques in object detection.
\newblock In \emph{2019 International Conference on Robotics and Automation
  (ICRA)}, pp.\  2348--2354. IEEE, 2019.

\bibitem[Oliphant(2006)]{oliphant2006guide}
Oliphant, T.~E.
\newblock \emph{A guide to NumPy}, volume~1.
\newblock Trelgol Publishing USA, 2006.

\bibitem[Paszke et~al.(2017)Paszke, Gross, Chintala, Chanan, Yang, DeVito, Lin,
  Desmaison, Antiga, and Lerer]{paszke2017automatic}
Paszke, A., Gross, S., Chintala, S., Chanan, G., Yang, E., DeVito, Z., Lin, Z.,
  Desmaison, A., Antiga, L., and Lerer, A.
\newblock Automatic differentiation in pytorch.
\newblock 2017.

\bibitem[Peng et~al.(2017)Peng, Luo, and Ni]{peng2017acts}
Peng, F., Luo, Q., and Ni, L.~M.
\newblock Acts: An active learning method for time series classification.
\newblock In \emph{2017 IEEE 33rd International Conference on Data Engineering
  (ICDE)}, pp.\  175--178. IEEE, 2017.

\bibitem[Scalia et~al.(2019)Scalia, Grambow, Pernici, Li, and
  Green]{scalia2019evaluating}
Scalia, G., Grambow, C.~A., Pernici, B., Li, Y.-P., and Green, W.~H.
\newblock Evaluating scalable uncertainty estimation methods for dnn-based
  molecular property prediction.
\newblock \emph{arXiv preprint arXiv:1910.03127}, 2019.

\bibitem[Siddhant \& Lipton(2018)Siddhant and Lipton]{siddhant2018deep}
Siddhant, A. and Lipton, Z.~C.
\newblock Deep bayesian active learning for natural language processing:
  Results of a large-scale empirical study.
\newblock \emph{arXiv preprint arXiv:1808.05697}, 2018.

\bibitem[Tsymbalov et~al.(2019)Tsymbalov, Makarychev, Shapeev, and
  Panov]{tsymbalov2019deeper}
Tsymbalov, E., Makarychev, S., Shapeev, A., and Panov, M.
\newblock Deeper connections between neural networks and gaussian processes
  speed-up active learning.
\newblock \emph{arXiv preprint arXiv:1902.10350}, 2019.

\bibitem[Wilson \& Izmailov(2020)Wilson and Izmailov]{wilson2020bayesian}
Wilson, A.~G. and Izmailov, P.
\newblock Bayesian deep learning and a probabilistic perspective of
  generalization.
\newblock \emph{arXiv preprint arXiv:2002.08791}, 2020.

\bibitem[Wilson et~al.(2015)Wilson, Hu, Salakhutdinov, and
  Xing]{Wilson2015DeepKL}
Wilson, A.~G., Hu, Z., Salakhutdinov, R., and Xing, E.~P.
\newblock Deep kernel learning.
\newblock In \emph{AISTATS}, 2015.

\bibitem[Zhang et~al.(2015)Zhang, Zou, He, and Sun]{zhang2015accelerating}
Zhang, X., Zou, J., He, K., and Sun, J.
\newblock Accelerating very deep convolutional networks for classification and
  detection.
\newblock \emph{IEEE transactions on pattern analysis and machine
  intelligence}, 38\penalty0 (10):\penalty0 1943--1955, 2015.

\end{thebibliography}
\bibliographystyle{icml2020}
\appendix

\twocolumn[
\icmltitle{Supplementary Material}]

\section{Implementation details}
Our methodology is as follows. We train a VGG-16 \citep{zhang2015accelerating} pretrained on ImageNet~\citep{imagenet_cvpr09}. Our initial training set contains 500 samples. We estimate the uncertainty using 20 MC samples and label the 100 most uncertain elements. Following \citet{gal2017deep}, we reset the weights to their initial value between steps. 

\section{Imbalanced datasets}
 
How to deal with imbalanced datasets is an entire area of research \citep{krawczyk2016learning}, but little has been done to deal with it when we are not aware of the \textit{a priori} class distribution. In consequence, the active learning model may quickly overfit to the more popular classes and reduce the effectiveness of active learning procedure. From \citet{gal2017deep}, it is known that Bayesian active learning will favor underrepresented classes. But, we find the reported experiments to be too simple. We  test this hypothesis in a controlled environment where we can set the number of unrepresented classes. 

In \tabref{fig:imbalanced}, we took the standard CIFAR100 dataset and we mimic an imbalanced dataset where few classes have a high number of examples. A class selected to be underrepresented sees its number of samples to be reduced by 75\%. When we increase the number of underrepresented classes, the gain of using MC-Dropout versus random sampling becomes more obvious. This is due to regions on the learned manifold associated with underrepresented classes to be highly uncertain. In consequence, these regions will be selected for labelling very early in the process.


\begin{table}
    \centering
    \begin{tabular}{llll}
    
    \toprule
    Dataset size &         5000 &        10000 &        20000 \\
    
    \hline
   $\Delta =10$ \\
    BALD &  4.39 \std{0.4} &  3.99 \std{0.01} &  3.57 \std{0.05} \\
    Entropy &  4.71 \std{0.02} &  4.54 \std{0.07} &  3.94 \std{0.01} \\
    Random &  4.52 \std{0.09} &  4.10 \std{0.03} &  3.71 \std{0.05} \\
    \hline
   $\Delta =25$ \\
    BALD &  4.40 \std{0.03}	&4.04\std{0.03} &	3.61\std{0.08} \\
    Entropy &  4.76 \std{0.02} &  4.68 \std{0.08} &   4.00 \std{0.01} \\
    Random &  4.58 \std{0.08} &  4.18 \std{0.04} &  3.75 \std{0.01} \\
    \hline 
   $\Delta =50$ \\
    BALD &  4.49 \std{0.08} &   4.07 \std{0.02} &   3.66 \std{0.04} \\
    Entropy &  4.83 \std{0.04} &  4.60 \std{0.14} &  4.07 \std{0.28} \\
    Random &  4.62 \std{0.03} &   4.21 \std{0.02} &   3.76 \std{0.04} \\
    \end{tabular}
    \caption{Effect of using active learning on imbalanced versions of CIFAR100. $\Delta$ is the number of class that contains 25\% of their data. From \citep{gal2017deep}, we know that BALD is robust to imbalanced datasets, but the study was not extensive. While BALD is robust to imbalanced datasets, the effect is catastrophic when using Entropy.  Performance averaged over 5 runs.}
    \label{fig:imbalanced}
\end{table}

\section{Effect of convergence}

\begin{figure}
    \centering
    \includegraphics[width=0.5\textwidth]{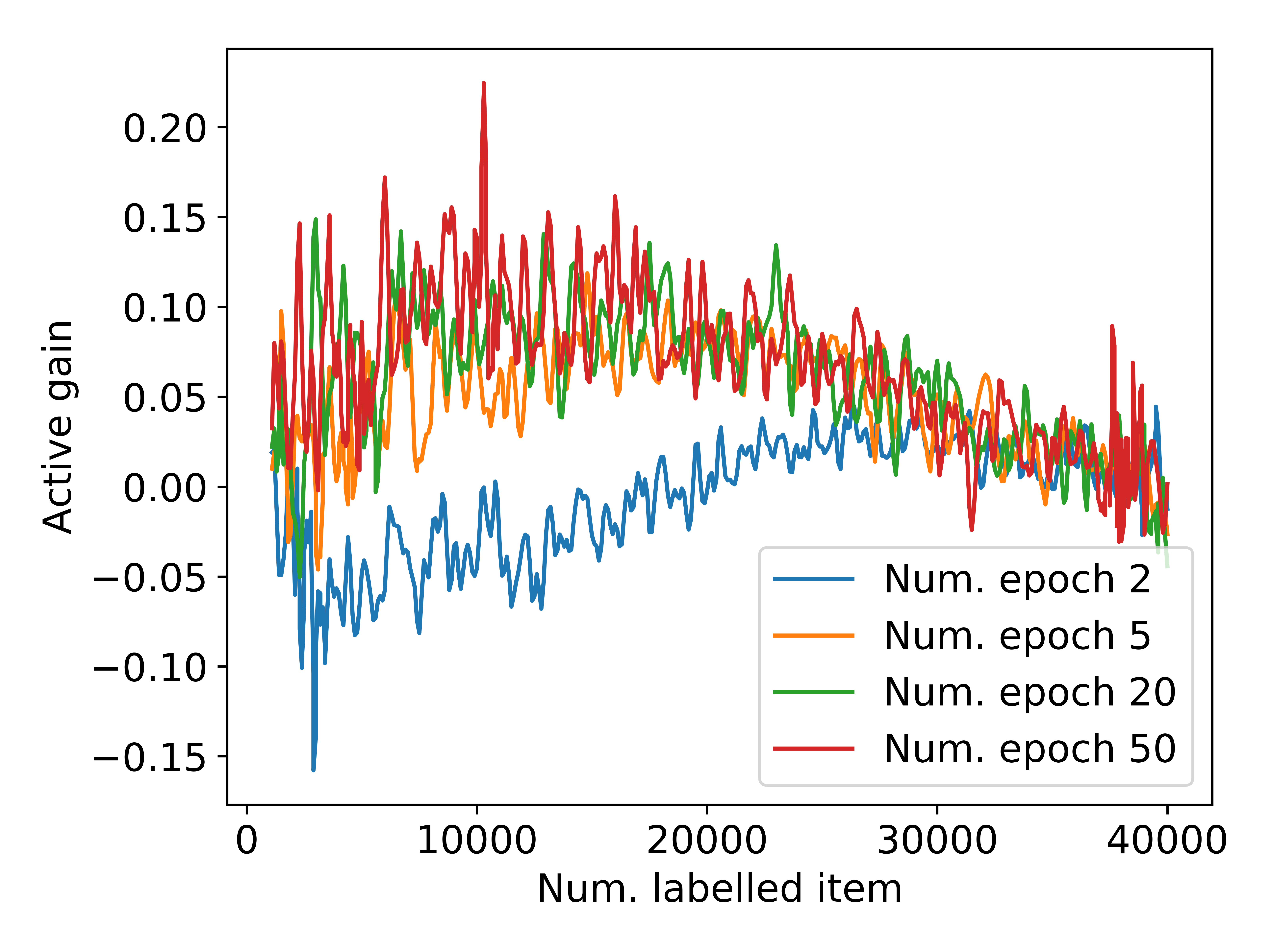}
    \caption{Gain of using active learning when varying the number of training epochs. An underfitted model will cause harm to the model training and in this case, just using random would've been better.}
    \label{fig:active_gain}
\end{figure}

In \figref{fig:active_gain}, we computed the difference in performance between BALD and random. We call this measure the \textit{Active gain} $= NLL_{Random} - NLL_{BALD}$. When using an underfitted model, the gain goes negative i.e. you would be better to use random selection.

\section{Effect of reducing the pool size}
\begin{figure}
    \centering
    \includegraphics[width=0.4\textwidth]{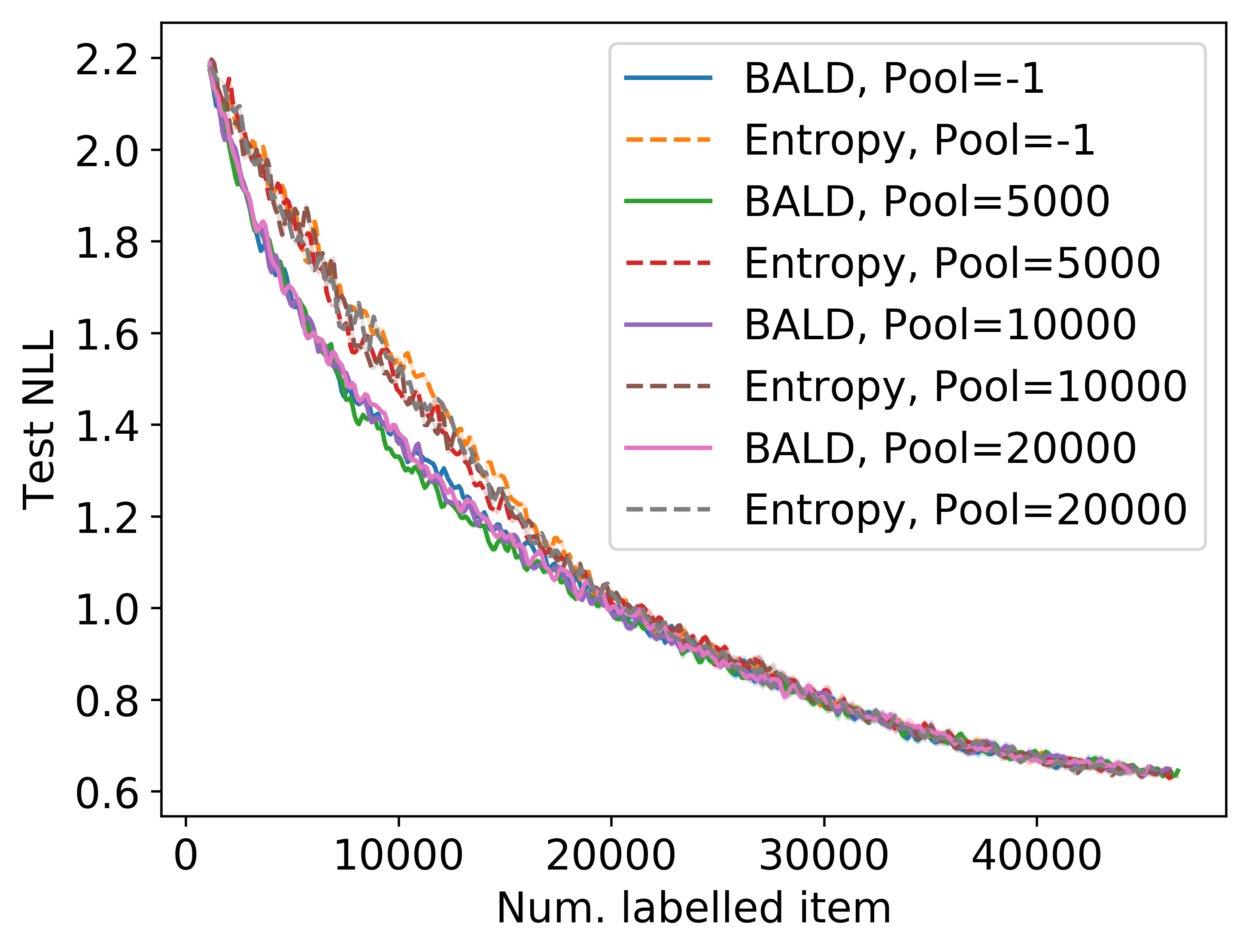}
    \caption{Effect of reducing the size of the pool on CIFAR100. \textbf{-1} indicates no reduction. For all heuristics, the performance is not affected by the size of the pool showing that AL can be efficient when tuned properly.  Performance averaged over 5 runs.}
    \label{fig:pool_size}
\end{figure}

As part of the experiments, we test whether limiting the pool size would affect the performance of active learning. Our experiments in  \figref{fig:pool_size} show that whether to calculate the uncertainty for the whole pool data or a randomly selected subset, the performance of active learning is not affected. This leads to an the interesting outcome of limiting the uncertainty calculations (which is the most expensive part of an active learning loop) in production setup for faster active learning loops.

\begin{figure}
    \centering
    \includegraphics{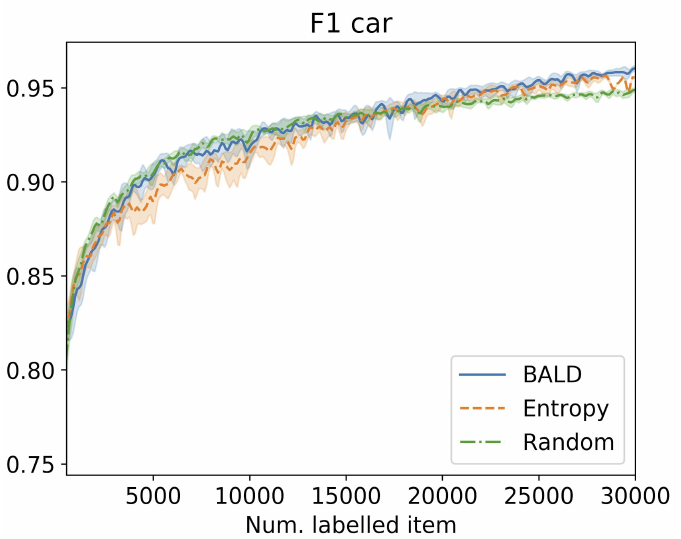}
    \caption{\textbf{F1 for the class \textit{car}}. BALD is great for underrepresented classes while not affecting more popular classes. Entropy decreases the performance on this class.}
    \label{fig:miotcd}
\end{figure}

\section{Bayesian active learning Library (BaaL)}

All the experiments in this paper have done using our publicly available Bayesian active learning library.  The goal of this library is to provide an easy to use but complete setup to test active learning on any project with few lines of code. We included features that current active learning libraries do not support. In particular, Bayesian methods such as MC-Dropout or Coresets are not widely available and there is no standard implementation of the active learning loop. Furthermore, research codebases are often hard to read and hard to maintain. Our proposed unified API could satisfy both research and industrial users.

 Our recently published open-source package named BaaL, aims at accelerating the transition from research to production. The core philosophy behind our library is to provide researchers with a well-designed API so that they focus on their novel idea and not on technical details. Our library proposes a task-agnostic system where one can mix-and-match any set of acquisition functions and uncertainty estimation methods.
The library consists of three main components:
\begin{enumerate}
    \item Dataset management to keep track and manage the labelled data $D_L$ and the unlabelled data $D_U$.
    \item Bayesian Methods i.e. MC-Dropout, MC-DropConnect and so on.
    \item Acquisition functions i.e. BALD, BatchBALD, Entropy and more.
\end{enumerate}

We provide full support for Pytorch \cite{paszke2017automatic} deep learning modules but our acquisition functions which are the most important part of active learning is implemented in  Numpy\cite{oliphant2006guide} and hence can be used on any platform. Our Data management module keeps track of what is labelled and what is unlabelled. We also provide facilitator methods to label a data point, update the pool of unlabelled data, and to randomly label a portion of the dataset. In our Bayesian module, we provide utilities to make any Pytorch model Bayesian with a single instruction. We also provide training, testing, and active learning loops that facilitate the active training procedure. Our acquisition functions are up-to-date with state-of-the-art methods. We provide easy to follow tutorials (https://baal.readthedocs.io/en/latest/) for each section of the library so that the user understands how each component works. Finally, our library is a member of Pytorch Ecosystem, which is reserved for libraries with outstanding documentation.

Our road-map has been indicated in the repository. Our current focus will include model calibration and semi-supervised learning. As more researchers contribute their methods to our library, we aim to become the standard Bayesian active learning library.

\end{document}